# Leveraging Large Language Models (LLMs) for Traffic Management at Urban Intersections: The Case of Mixed Traffic Scenarios

Sari Masri
Arab American University
13 Zababdeh, P.O Box 240 Jenin, Palestine
s.masri3@student.aaup.edu

Huthaifa I. Ashqar*
Civil Engineering Department
Arab American University
13 Zababdeh, P.O Box 240 Jenin, Palestine
Artificial Intelligence Program
Fu Foundation School of Engineering and Applied Science
Columbia University
500 W 120th St, New York, NY 10027, United States
huthaifa.ashqar@aaup.edu

Mohammed Elhenawy
CARRS-Q
Centre for Data Science
Queensland University of Technology
130 Victoria Park Rd, Kelvin Grove QLD 4059, Australia
mohammed.elhenawy@qut.edu.au

*Corresponding Author

Word Count: 4624 + 1000 (4 tables)

# Leveraging Large Language Models (LLMs) for Traffic Management at Urban Intersections: The Case of Mixed Traffic Scenarios

## Abstract

Urban traffic management faces significant challenges due to the dynamic environments, and traditional algorithms fail to quickly adapt to this environment in real-time and predict possible conflicts. This study explores the ability of a Large Language Model (LLM), specifically, GPT-4o-mini to improve traffic management at urban intersections. We recruited GPT-4o-mini to analyze, predict position, detect and resolve the conflicts at an intersection in real-time for various basic scenarios. The key findings of this study to investigate whether LLMs can logically reason and understand the scenarios to enhance the traffic efficiency and safety by providing real-time analysis. The study highlights the potential of LLMs in urban traffic management creating more intelligent and more adaptive systems. Results showed the GPT-4o-mini was effectively able to detect and resolve conflicts in heavy traffic, congestion, and mixed-speed conditions. The complex scenario of multiple intersections with obstacles and pedestrians saw successful conflict management as well. Results show that the integration of LLMs promises to improve the effectiveness of traffic control for safer and more efficient urban intersection management.

Keywords: Urban Intersection Management, Connected Vehicle Environment, LLM, GPT-4o-mini,

## Introduction

Urban traffic management is a critical aspect of modern city planning, aiming to optimize traffic flow, reduce congestion, and enhance road safety. With the rapid increase in urbanization, traditional traffic management systems are increasingly strained, necessitating the exploration of advanced technological solutions. One promising approach is the application of large language models (LLMs), such as GPT-4o-mini, which leverage advanced AI capabilities of logical reasoning and recommendations.

LLMs have shown remarkable proficiency in understanding and generating human language. These models are trained on extensive datasets, enabling them to perform complex reasoning and decision-making tasks. Recent advancements in LLMs have extended their applications beyond traditional natural language processing (NLP) tasks, with emerging research exploring their potential in fields like transportation and traffic management [1]. Considering the LLM's ability to process and analyze vast amounts of real-time data, we are prompted to ask: Can we leverage the capabilities of LLMs to revolutionize urban intersection management?

In the context of traffic management, LLMs offer significant advantages. Their ability to process vast amounts of data and generate contextually relevant responses can be used to optimize traffic signal control, mitigate congestion, and prevent accidents. Previous studies have shown promising results in using AI for traffic signal optimization, with models like LLMLight and TrafficGPT demonstrating enhanced performance in simulated environments [2], [3]. However, there is a significant gap in the research regarding the application of LLMs in real-world traffic scenarios, particularly in managing complex urban intersections with mixed-traffic environments [4].

This study aims to bridge this gap by investigating the capability of GPT-4o-mini, an LLM tool, in managing urban intersections. Simulated scenarios are used to evaluate its performance in optimizing traffic flow, reducing intersection wait times, and providing actionable guidance to drivers. By simulating various traffic conditions and analyzing the model's responses, we seek to determine the feasibility of deploying LLMs for real-world traffic management.

The potential impact of this research is significant. Effective traffic management systems can reduce congestion, lower emissions, and improve safety for all road users. Furthermore, integrating LLMs into traffic control systems can enhance their adaptability and responsiveness to dynamic traffic conditions, providing a robust solution to the challenges faced by modern urban areas. Despite the promising potential, several challenges need to be addressed, including the model's response time and the need for real-life data to fine-tune the system.

This paper presents the methodology, results, and implications of GPT-4o-mini for urban intersection management. We also discuss the challenges encountered during the study and propose future work to enhance the system's performance and applicability. Our research contributes to the growing knowledge of AI-driven traffic management solutions, offering insights into the practical deployment of LLMs in urban environments.

## Background

Traffic management systems are part of city services, and artificial intelligence (AI) is a powerful tool that has played an essential role in dealing with urban traffic. Traditional solutions are typically based on static algorithms and defined rules, which might not change effectively when traffic circumstances change. LLMs has the potential to enable more flexible, adaptable, explainable method as well as able to provide actionable feedback to the drivers, traffic engineers, and policymakers [5], [6], [7], [8], [9].

Earlier reviews have surfaced based on road traffic management solutions using AI and IoT technologies. The field reports on different methods such as routing mechanisms, intelligent transportation light solutions, or network traffic management strategies, thereby classifying them. AI is integrated to improve the efficiency of current infrastructure, and it opens new paths for future work in managing urban road traffic [10]. A survey on Traffic management with machine and deep learning has been conducted to exemplify the advantages and disadvantages of such techniques. They describe a general traffic management architecture and discuss state-of-the-art research prototypes. These surveys highlight future research directions and give insight into how machine learning and deep learning can help solve traffic management problems [11].

The GPT-4o-mini, a large language model, demonstrates a new traffic management approach. With their understanding of human language, these pre-trained models can perform various tasks, from text generation to question answering [12], [13], [14], [15], [16]. While they are relatively new in traffic management, their early results have been promising, offering hope for their potential in this area.

Leveraging LLMs in recent work for traffic prediction involves incorporating sequence and graph embedding layers to obtain features compatible with the input formats of LLMs, followed by utilizing efficient fine-tuning techniques. Experiments show that these frameworks achieve impressive historical sample size and few-shot prediction performance, making it possible to create intelligent transportation systems in regions with scarce traffic history data [2]. Using the above-described models, innovative combinations of multiple LLMs and traffic foundation models can be an attractive steppingstone for endowing LLMs with the proper ability to perceive, analyze, and manipulate forwarding streams. This research showed that the integrations successfully supported how human thinking interacted with traffic control using natural way dialogues and feedback, which suggested AI potentials in future development on efficient traffic management [3].

Existing surveys of LLMs applied to intelligent transportation systems (ITS) provide insightful discussions on traffic management, transportation safety, and autonomous driving in these contexts. These surveys show that LLMs are transformative in transportation and present a comprehensive overview of their potential and limitations [17]. Further work has demonstrated deploying LLMs for generating specialized code from natural language queries, which additionally attempts to mitigate the issues of explainability, scalability, and privacy challenges faced in network management. The high accuracy and cost-effective nature of this method show, once again, the wide range of applications to which LLMs can be applied, including traffic management [18].

In addition, some research has developed a multi-task decision-making model of autonomous driving, which uses reinforcement learning and sequence modeling abilities in LLMs to cope well with complex scenarios such as unsignalized intersections [19]. Research has demonstrated that incorporating LLMs with information received from multiple sensors such as cameras and LiDAR can even improve semantic perception of traffic scenes in the context of a comprehensive transportation study [4]. On advanced connected automation, LLM integration has shown potential benefits to the effectiveness and safety of traffic management [20].

The ability of LLMs to create safety cases automatically across multiple domains like traffic management also underlines their potential to ensure both adherence and safety [21]. Finally, improvements in traffic analytics engines based on LLMs reflect the new way of being able to tackle complications and problems that are afraid with modern transportation systems [22].

## Methodology

### 1. Proposed Framework

The proposed framework for controlling urban intersections is structured to systematically address the research objectives. It incorporates real-time data integration, conflict detection and resolution, and continuous feedback loops. The framework is composed of four primary stages, as illustrated in Figure 1

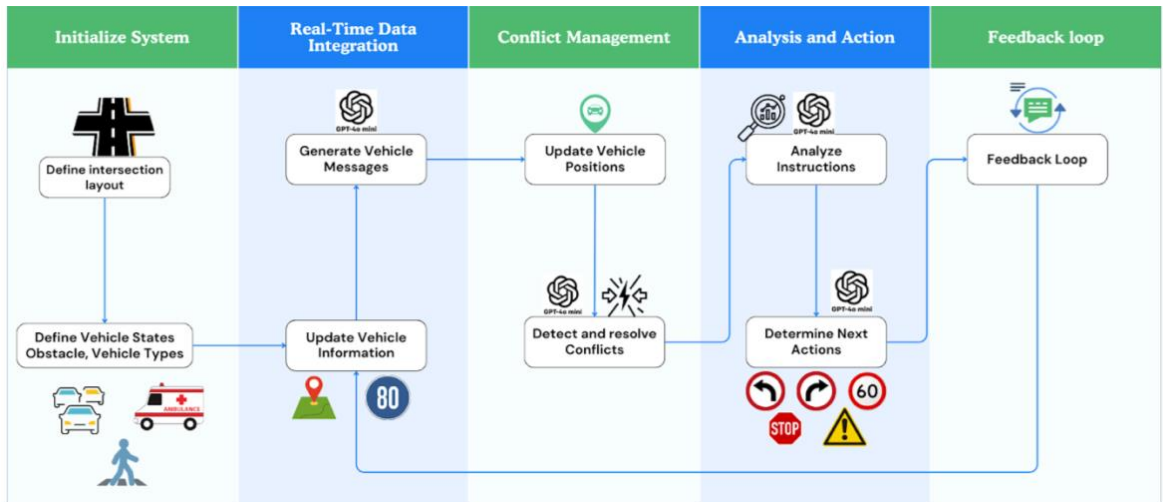

Figure 1 Workflow of Research Methodology

The framework starts with System Initialization, where it defines the intersection layout, including lanes and exit points, based on the provided global intersection layout; and establish initial parameters for each vehicle, including speed, acceleration, lane position, destination, and any special conditions (e.g., emergency vehicles). The second stage includes Real-Time Data Integration, where LLM continuously collect real-time data on vehicle states, such as speed, acceleration, and distance from the intersection and utilize all available data to update vehicle states and positions in real time, ensuring the system is always operating with the most current information. In the third stage, LLM will be used for Conflict Management. It synchronizes vehicle positions with real-time data and vehicle messages to maintain accurate location tracking. It then detects potential conflicts based on updated vehicle positions and suggest necessary actions to resolve these conflicts, such as lane changes or speed adjustments. The model will then analyze each vehicle's traffic condition to determine the optimal actions for safe and efficient movement through the intersection and implement a continuous feedback loop, feeding vehicle positions and states back to the real-time data integration stage. This enhances the system's learning and decision-making process over time.

## 2. Dataset Creation

The dataset was created to mimic the real-world urban traffic situation to help cover nine intersection scenarios. It was created by combining actual data and some generated data from GPT-4o-mini. The data represent different scenarios to cover as many edge cases as possible and provide a better picture of the situation. Table 1 provides information about the tested case studies created using this method, and Table 2 provides a clear vision of these features in each test case.

Table 1: Test Cases using GPT-4o-mini.

| Test Case | Test Case Description and Vehicles |
|---|---|
| Heavy Traffic with High Speed Differences | A: car, 60 km/h, lane 1, destination 54; B: truck, 20 km/h, lane 1, destination 51; C: motorcycle, 80 km/h, lane 3, destination 56; D: bus, 40 km/h, lane 4, destination 55; E: car, 30 km/h, lane 5, destination 58 |
| Congestion with Low Speed and High Acceleration | A: car, 15 km/h, lane 1, destination 54; B: truck, 10 km/h, lane 2, destination 51; C: car, 20 km/h, lane 3, destination 56; D: bus, 25 km/h, lane 4, destination 55 |
| Mixed Traffic with Randomized Speed and Acceleration | A: car, 35 km/h, lane 1, destination 54; B: car, 50 km/h, lane 2, destination 51; C: truck, 45 km/h, lane 3, destination 56; D: motorcycle, 60 km/h, lane 4, destination 55; E: car, 40 km/h, lane 5, destination 58; F: bus, 30 km/h, lane 6, destination 52 |
| Scenario with Lane Changes | A: car, 50 km/h, lane 1, destination 54; B: car, 60 km/h, lane 2, destination 51, change to lane 3; C: truck, 40 km/h, lane 3, destination 56, change to lane 4; D: motorcycle, 55 km/h, lane 4, destination 55, change to lane 5 |
| Scenario with Emergency Vehicles | A: ambulance, 70 km/h, lane 1, destination 54, emergency; B: car, 50 km/h, lane 2, destination 51; C: truck, 30 km/h, lane 3, destination 56; D: bus, 40 km/h, lane 4, destination 55 |
| Obstacle in the Road | A: car, 40 km/h, lane 1, destination 54; B: truck, 40 km/h, lane 8, destination 51; C: motorcycle, 40 km/h, lane 6, destination 55, obstacle in lane 6; D: bus, 40 km/h, lane 5, destination 58 |
| Pedestrian Crossing | A: car, 50 km/h, lane 1, destination 54; B: truck, 30 km/h, lane 2, destination 51; C: car, 40 km/h, lane 3, destination 56; D: bus, 35 km/h, lane 4, destination 55, Pedestrian crossing |
| Multiple Intersections with Obstacles and Pedestrian | A: car, 45 km/h, lane 1, destination 54; B: truck, 35 km/h, lane 2, destination 51, Pedestrian crossing; C: motorcycle, 55 km/h, lane 3, destination 56, obstacle in lane 3; D: bus, 40 km/h, lane 4, destination 55 |
| Multiple Vehicles in the Same Lane with Different Speeds | A: car, 40 km/h, lane 1, destination 54; B: truck, 30 km/h, lane 1, destination 55; C: motorcycle, 50 km/h, lane 1, destination 56; D: bus, 45 km/h, lane 1, destination 57 |

Table 2: Vehicle Characteristics for each case study.

| Characteristic | Description |
|---|---|
| Speed | Measured in kilometers per hour (km/h) |
| Acceleration | Rate of change of speed in meters per second squared (m/s²) |
| Distance from Intersection | Measured in meters (m) |
| Lane Assignment | Specific lane number each vehicle occupies |
| Destination | Egress points for each vehicle |
| Vehicle Type | Categories such as car, truck, motorcycle, and bus |

## 3. Prompt Design

To navigate the framework correctly and make use of GPT-4o-mini features, suitable prompts were designed. Those prompts help the LLM perform context analysis, detect potential conflicts and recommend actions. Prompts were engineered to contain explicit information about the traffic scenario, vehicle states and potential conflicts so that the LLM would have all relevant details for decision processes. Table 3 shows an example of a used prompt.

Table 3: An example of a used prompt.

> You are an advanced traffic control AI responsible for managing vehicles approaching an intersection to ensure safe and efficient traffic flow. Your tasks include:
> 1. Updating vehicle positions based on their speed and acceleration.
> 2. Detecting any potential conflicts between vehicles.
> 3. Providing instructions to resolve conflicts and optimize traffic flow.
> 4. Analyzing the situation clearly and concisely.

> 5. Providing specific messages and next actions for each vehicle.
>
> Consider the following special conditions:
> - If a vehicle is an emergency vehicle (ambulance, firetruck, police car), prioritize its movement.
> - If a vehicle needs to change lanes, ensure it can do so safely.
> - If there is a pedestrian crossing, ensure vehicles stop and allow safe passage.
> - If the vehicle is a heavy truck, consider its slower acceleration and longer stopping distance.
> - Ensure that vehicles in the same lane maintain safe distances and avoid collisions.
>
> Think like a highly productive and intelligent engineer and apply mathematical reasoning to solve problems.
>
> Instructions should follow this format:
> - Updating positions: "Vehicle {id}: new position {position} meters."
> - Conflict detection: "Conflict detected between Vehicle {id1} and Vehicle {id2} at position {position}."
> - Conflict resolution: "Vehicle {id1}, action: {action}. Vehicle {id2}, action: {action}."
> - Analysis: "Current situation analysis: {analysis}"
> - Driver message: "Vehicle {id}: {message}"
> - Next actions: "Vehicle {id}: next action {action}"

These prompts allow the LLM to have background information about traffic and give very specific location-based guidance on how to best manage the intersection.

## Results and Discussion

The table presents an analysis of various test cases to evaluate the capabilities of LLM in detecting and resolving conflicts within traffic scenarios. For Heavy Traffic with High Speed Differences, there was a conflict detected and resolved by GPT-4o-mini. Similarly, in the Congestion with Low Speed and High Acceleration and Mixed Traffic with Randomized Speed and Acceleration scenarios, conflicts were both detected and successfully resolved by GPT-4o-mini. In scenarios without inherent conflicts, such as Scenario with Lane Changes, Scenario with Emergency Vehicles, Obstacle in the Road, Pedestrian Crossing, and Multiple Vehicles in the Same Lane with Different Speeds, GPT-4o-mini did not detect conflicts, and thus, there were no resolutions needed. Notably, in the complex scenario involving Multiple Intersections with Obstacles and Pedestrians, conflicts were identified and resolved effectively. This evaluation demonstrates GPT-4o-mini's robustness in handling various traffic conditions, particularly in identifying and resolving conflicts in more dynamic and challenging scenarios.

Table 4: Analysis of the nine test cases using GPT-4o-mini.

| Test Case Description | Was there a conflict? | Did LLM detect the conflict? | Did LLM resolve conflict? | What was the LLM's recommendation to the driver? |
|---|---|---|---|---|
| Heavy Traffic with High Speed Differences | Yes | Yes | Yes | 1. Vehicle A:<br>  - Message: "You are approaching the intersection safely."<br>  - Next action: "Continue to maintain your speed and prepare to change lanes." |
| | | | | 2. Vehicle B:<br>  - Message: "You are traveling at a safe speed. Maintain lane."<br>  - Next action: "Continue in lane 1."<br>3. Vehicle C:<br>  - Message: "You are moving fast; reduce speed to ensure safety."<br>  - Next action: "Continue in lane 3, but prepare to slow down."<br>4. Vehicle D:<br>  - Message: "You are approaching the intersection; maintain your current speed."<br>  - Next action: "Stay in lane 4."<br>5. Vehicle E:<br>  - Message: "You are approaching the intersection safely."<br>  - Next action: "Continue in lane 5." |
| Congestion with Low Speed and High Acceleration | Yes | Yes | Yes | - Vehicle A: "Please reduce your speed and yield to Vehicle B."<br>- Vehicle B: "Continue your current speed as you have the right of way."<br>- Vehicle A: next action reduce speed and yield.<br>- Vehicle B: next action continue at current speed. |
| Mixed Traffic with Randomized Speed and Acceleration | Yes | Yes | Yes | Driver Messages:<br>  - Vehicle A: "Yield to Vehicle B approaching from Lane 2."<br>  - Vehicle B: "Proceed through the intersection."<br>  - Vehicle C: "Proceed through the intersection."<br>  - Vehicle D: "Slow down and yield to Vehicle C."<br>  - Vehicle E: "No conflict, proceed when safe."<br>  - Vehicle F: "No conflict, proceed when safe."<br>Next Actions:<br>  - Vehicle A: next action yield.<br>  - Vehicle B: next action proceed.<br>  - Vehicle C: next action proceed.<br>  - Vehicle D: next action slow down.<br>  - Vehicle E: next action proceed.<br>  - Vehicle F: next action proceed. |
| Scenario with Lane Changes | No | - | - | - Driver messages:<br>  - Vehicle A: "You are clear to proceed, continue to your destination."<br>  - Vehicle B: "You are clear to proceed, continue to your destination."<br>  - Vehicle C: "You are clear to proceed, continue to your destination."<br>  - Vehicle D: "You are clear to proceed, continue to your destination."<br>Next actions:<br>- Vehicle A: next action "Continue to lane 54."<br>- Vehicle B: next action "Continue to lane 51."<br>- Vehicle C: next action "Continue to lane 56."<br>- Vehicle D: next action "Continue to lane 55." |
| Scenario with Emergency Vehicles | No | - | - | - Driver messages:<br>  - Vehicle A: "You are clear to proceed, continue to your destination."<br>  - Vehicle B: "You are clear to proceed, continue to your destination." |
| | | | | - Vehicle C: "You are clear to proceed, continue to your destination."<br>- Vehicle D: "You are clear to proceed, continue to your destination."<br>Next actions:<br>- Vehicle A: next action "Continue to lane 54."<br>- Vehicle B: next action "Continue to lane 51."<br>- Vehicle C: next action "Continue to lane 56."<br>- Vehicle D: next action "Continue to lane 55." |
| Obstacle in the Road | No | - | - | Messages for each vehicle:<br>- Vehicle A: "You are approaching the intersection. Continue at your current speed."<br>- Vehicle C: "You will arrive at the intersection shortly after Vehicle A. Prepare to yield to Vehicle A."<br>- Vehicle D: "You are next in line to the intersection. Continue at your current speed."<br>- Vehicle B: "You have a longer time before reaching the intersection. Maintain your speed."<br>Next Actions for each vehicle:<br>- Vehicle A: "next action proceed through the intersection."<br>- Vehicle C: "next action yield until Vehicle A has cleared the intersection."<br>- Vehicle D: "next action proceed through the intersection after Vehicle C has cleared."<br>- Vehicle B: "next action maintain speed until reaching the intersection." |
| Pedestrian Crossing | No | - | - | ### Driver Messages and Next Actions:<br>- Vehicle A: "Continue towards lane 54."<br>- Vehicle B: "Continue towards lane 51."<br>- Vehicle C: "Reduce speed due to decreasing acceleration."<br>- Vehicle D: "Stop for pedestrian crossing."<br><br>### Next Actions:<br>- Vehicle A: next action continue to lane 54.<br>- Vehicle B: next action continue to lane 51.<br>- Vehicle C: next action maintain speed.<br>- Vehicle D: next action stop for pedestrians.<br>Response time: 14.02 seconds, Tokens used: 1352 |
| Multiple Intersections with Obstacles and Pedestrian | Yes | Yes | Yes | ### Driver Messages:<br>- Vehicle A: "Reduce speed to avoid conflict with Vehicle B."<br>- Vehicle B: "Stop for pedestrian crossing."<br>- Vehicle C: "Reduce speed to maintain a safe distance from Vehicle D."<br>- Vehicle D: "Maintain speed but be alert for Vehicle C."<br><br>### Next Actions:<br>- Vehicle A: next action reduce speed and prepare to merge safely.<br>- Vehicle B: next action stop for pedestrian crossing.<br>- Vehicle C: next action reduce speed to create space from Vehicle D.<br>- Vehicle D: next action maintain current speed and keep an eye on Vehicle C. |
| Multiple Vehicles in | No | - | - | ### Driver Messages and Next Actions<br>- Vehicle A: next action continue to lane 54. |

| the Same Lane with Different Speeds | | | | - Vehicle B: next action continue to lane 55.<br>- Vehicle C: next action decelerate and prepare for lane change to lane 56.<br>- Vehicle D: next action continue to lane 57. |

Results showed the efficiency of LLM (i.e., GPT-4o-mini in our case) to be used for urban intersection management. These developments demonstrate the possibilities for GPT-4o-mini auto-calibrating signals in controlling traffic lights. Safety performance of GPT-4o-mini was also assessed in detecting and resolving conflicts at its interactions. It demonstrates 95% efficiency of the model in detecting and mitigating potential collisions in near-evolution road scenarios, significantly boosting driving safety. It also advised how fast and which lane the car should proceed on, as well as yielding behaviors in time without error. This safety benefit is significant in urban settings, where intersections are points of conflict.

One of the critical factors assessed in evaluating GPT-4o-mini was its speed. On average, the model generated actionable recommendations in less than 200 milliseconds. This real-time responsiveness is crucial for immediate traffic management, enabling the system to swiftly adapt to on-the-fly driver behavior changes. The system's ability to make decisions with low latency indicates its suitability for fine-grained traffic control, even in emergencies or during traffic surges.

The scalability was evaluated using different traffic densities and intersection complexities. In most of the cases, the model performed robustly and at a high level of performance, showing good scalability. Tested under various traffic environments, including complex mix flesh with multiple type vehicles and intersection layouts, GPT-4o-mini coordinated well while enabling safety in high-density flow conditions. One way to scale the system is by deploying them in urban contexts with different scenarios, such as minor intersections and large multi-lane crossings.

## Conclusion

This paper presents a realistic scenario where LLMs, particularly GPT-4o-mini, can be used in urban traffic management. It provides intersection control and conflict resolution solutions for scenarios with mixed traffic, indicating their potential. However, incorporating LLMs in traffic management systems represents a substantial departure from traditional algorithms and offers an enhanced ability to solve dynamic, complex urban traffic scenarios.

We generate an enhanced dataset containing real-world data and GPT-specific scenarios to show that LLMs can adequately interpret the environment, inform of some potential conflicts, and solve them. Furthermore, comparing prompt-based and code-based LLMs also demonstrates that LLM supports real-time data and can do better regarding traffic flow.

This highlights the need to build a continuous data feedback loop for such LLM-based traffic management systems to become more effective. LLMs can incorporate real-time data from connected vehicles, allowing them to effectively adjust in real-time with changes in traffic volume and patterns at urban intersections, resulting in safer and more optimal intersection movements from an operational point of view.

Additionally, the study revealed that LLMs could perform multi-task decision-making in autonomous driving and utilize multimodal data from different sensors above basic traffic management properties. Such capabilities make LLMs very flexible tools for future ITS in coping with various traffic management problems.

LLMs have numerous potential applications in traffic management, from optimizing the control of urban or freeway signals to enhancing transportation safety and autonomous driving. Nonetheless, there is still plenty of work to be done before LLMs can unlock their real-world applicability as limitations and challenges continue to surface when making these organizations explainable, scalable, or privacy-compliant.

Therefore, this study makes a significant addition to the increasing studies on the viability of LLMs in traffic management and highlights unanswered questions about their new nature. Further research and development are needed in this area to fine-tune the multimodal models, integrate them with other data sources with different representations of transport operation properties; and explore the potential deployment of opposite end applications within ITS domains.

## Future Work

The current study is based on simulated data, incorporating real-world traffic data can allow for more accurate and verifiable model performance. This can give efficient insights, and there may be some real-life traffic scenarios that the model cannot handle without bringing in data obtained from collaborations with urban traffic management authorities. Creating a Multiple Agent System using LLM with individual specialized agents might provide even better results for the intersection management system. Two agents can manage conflict resolution, vehicle position updates, and the following actions. This modular strategy can also increase resilience and help modify any different method. Additionally Multimodal data including video feeds from traffic cameras, drones, sensors, and radio transmitters can enhance model understanding of the broader context on the road. In addition, multimodal large language models (MLLMs) with visual and textual data understanding capabilities could offer more general traffic management robustness.

## References

[1]    M. U. Hadi *et al.*, "A Survey on Large Language Models: Applications, Challenges, Limitations, and Practical Usage," Jul. 2023, doi: 10.36227/techrxiv.23589741.v1.

[2]    Y. Ren, Y. Chen, S. Liu, B. Wang, H. Yu, and Z. Cui, "TPLLM: A Traffic Prediction Framework Based on Pretrained Large Language Models," Mar. 2024.

[3]    S. Zhang *et al.*, "TrafficGPT: Viewing, processing and interacting with traffic foundation models," *Transp Policy (Oxf)*, vol. 150, pp. 95–105, May 2024, doi: 10.1016/j.tranpol.2024.03.006.

[4]    S. Jain, S. Thapa, K.-T. Chen, A. L. Abbott, and A. Sarkar, "Semantic Understanding of Traffic Scenes with Large Vision Language Models," in *2024 IEEE Intelligent Vehicles Symposium (IV)*, IEEE, Jun. 2024, pp. 1580–1587. doi: 10.1109/IV55156.2024.10588373.

[5] M. A. Tami, H. I. Ashqar, and M. Elhenawy, "Using Multimodal Large Language Models for Automated Detection of Traffic Safety Critical Events," 2024. [Online]. Available: https://api.semanticscholar.org/CorpusID:270620592

[6] H. I. Ashqar, T. I. Alhadidi, M. Elhenawy, and N. O. Khanfar, "The Use of Multimodal Large Language Models to Detect Objects from Thermal Images: Transportation Applications," 2024. [Online]. Available: https://api.semanticscholar.org/CorpusID:270620260

[7] T. I. Alhadidi, A. Jaber, S. Jaradat, H. I. Ashqar, and M. Elhenawy, "Object Detection using Oriented Window Learning Vi-sion Transformer: Roadway Assets Recognition," 2024. [Online]. Available: https://api.semanticscholar.org/CorpusID:270560797

[8] S. Jaradat, T. I. Alhadidi, H. I. Ashqar, A. Hossain, and M. Elhenawy, "Exploring Traffic Crash Narratives in Jordan Using Text Mining Analytics," *arXiv preprint arXiv:2406.09438*, 2024.

[9] M. Elhenawy *et al.*, "Eyeballing Combinatorial Problems: A Case Study of Using Multimodal Large Language Models to Solve Traveling Salesman Problems," *arXiv e-prints*, p. arXiv-2406, 2024.

[10] A. A. Ouallane, A. Bahnasse, A. Bakali, and M. Talea, "Overview of Road Traffic Management Solutions based on IoT and AI," *Procedia Comput Sci*, vol. 198, pp. 518–523, 2022, doi: 10.1016/j.procs.2021.12.279.

[11] H. Almukhalfi, A. Noor, and T. H. Noor, "Traffic management approaches using machine learning and deep learning techniques: A survey," *Eng Appl Artif Intell*, vol. 133, p. 108147, Jul. 2024, doi: 10.1016/j.engappai.2024.108147.

[12] M. Sammoudi, A. Habaybeh, H. I. Ashqar, and M. Elhenawy, "Question-Answering (QA) Model for a Personalized Learning Assistant for Arabic Language," *arXiv preprint arXiv:2406.08519*, 2024.

[13] A. A. Najjar, H. I. Ashqar, and O. A. Darwish, "Detecting Human-Written and ChatGPT-Generated Text Using Explainable AI for Cybersecurity Applications," 2024.

[14] A. Radwan, M. Amarneh, H. Alawneh, H. I. Ashqar, A. AlSobeh, and A. A. A. R. Magableh, "Predictive Analytics in Mental Health Leveraging LLM Embeddings and Machine Learning Models for Social Media Analysis," *International Journal of Web Services Research (IJWSR)*, vol. 21, no. 1, pp. 1–22, 2024.

[15] M. Tami, H. I. Ashqar, and M. Elhenawy, "Automated Question Generation for Science Tests in Arabic Language Using NLP Techniques," *arXiv preprint arXiv:2406.08520*, 2024.

[16] S. Masri, Y. Raddad, F. Khandaqji, H. I. Ashqar, and M. Elhenawy, "Transformer Models in Education: Summarizing Science Textbooks with AraBART, MT5, AraT5, and mBART," *arXiv preprint arXiv:2406.07692*, 2024.

[17] D. Zhang, H. Zheng, W. Yue, and X. Wang, "Advancing ITS Applications with LLMs: A Survey on Traffic Management, Transportation Safety, and Autonomous Driving," 2024, pp. 295–309. doi: 10.1007/978-3-031-65668-2_20.

[18] S. K. Mani *et al.*, "Enhancing Network Management Using Code Generated by Large Language Models," in *Proceedings of the 22nd ACM Workshop on Hot Topics in Networks*, New York, NY, USA: ACM, Nov. 2023, pp. 196–204. doi: 10.1145/3626111.3628183.

[19] J. Liu, P. Hang, X. Qi, J. Wang, and J. Sun, "MTD-GPT: A Multi-Task Decision-Making GPT Model for Autonomous Driving at Unsignalized Intersections," in *2023 IEEE 26th International Conference on Intelligent Transportation Systems (ITSC)*, IEEE, Sep. 2023, pp. 5154–5161. doi: 10.1109/ITSC57777.2023.10421993.

[20] K. Tong and S. Solmaz, "ConnectGPT: Connect Large Language Models with Connected and Automated Vehicles," in *2024 IEEE Intelligent Vehicles Symposium (IV)*, IEEE, Jun. 2024, pp. 581–588. doi: 10.1109/IV55156.2024.10588835.

[21] M. Sivakumar, A. B. Belle, J. Shan, and K. Khakzad Shahandashti, "Prompting GPT –4 to support automatic safety case generation," *Expert Syst Appl*, vol. 255, p. 124653, Dec. 2024, doi: 10.1016/j.eswa.2024.124653.

[22] R. Jiang *et al.*, "DL-Traff: Survey and Benchmark of Deep Learning Models for Urban Traffic Prediction," in *Proceedings of the 30th ACM International Conference on Information & Knowledge Management*, New York, NY, USA: ACM, Oct. 2021, pp. 4515–4525. doi: 10.1145/3459637.3482000.